\documentclass[11pt,a4paper]{article}
\usepackage[usenames, dvipsnames]{xcolor}
\usepackage[hyperref]{acl2019}
\usepackage{times}
\usepackage{latexsym}
\usepackage{url}
\usepackage[utf8x]{inputenc}
\usepackage{caption}
\usepackage{float}
\usepackage{graphicx}
\usepackage[hang,flushmargin]{footmisc}
\usepackage{booktabs}
\usepackage{stmaryrd}

\aclfinalcopy 
\def\aclpaperid{1832} 

\title{Exploring Phoneme-Level Speech Representations for \\ End-to-End Speech Translation}

\author{Elizabeth Salesky$^1$, Matthias Sperber$^2$, Alan W Black$^1$ \\
$^1$Carnegie Mellon University, USA \\
$^2$Karlsruhe Institute of Technology, Germany \\
\texttt{esalesky@cs.cmu.edu}}

\begin{document}
\maketitle

\begin{abstract}
Previous work on end-to-end translation from speech has primarily used frame-level features as speech representations, which creates longer, sparser sequences than text. 
We show that a na{\"i}ve method to create compressed phoneme-like speech representations is far more effective and efficient for translation than traditional frame-level speech features. 
Specifically, we generate phoneme labels for speech frames and average consecutive frames with the same label to create shorter, higher-level source sequences for translation.
We see improvements of up to 5 BLEU on both our high and low resource language pairs,  with a reduction in training time of 60\%.
Our improvements hold across multiple data sizes and two language pairs. 
\end{abstract}

\section{Introduction}

The way translation input is represented has been shown to impact performance as well as how much data the model requires to train \cite{sennrich2016bpeacl,salesky2018segmentation,cherry2018emnlp}.
The current standard approach for text-based translation is to segment words into subword units as a preprocessing step \cite{sennrich2016bpeacl}. Clustering common character sequences increases frequency of units in data and improves generalization to new word forms and rare words.
End-to-end speech-to-text models are showing competitive results \cite{weiss2017sequence,bansal2018low,bansal2018pre,berard2018end,anastasopoulos2018tied}, but so far have not compared different ways to represent speech input.
Unlike text, where discrete trainable embeddings are typically used, speech models typically use continuous features extracted from sliding windows (frames), held fixed during training. 
Frame-level features yield significantly longer, more sparsely-represented sequences than their text equivalents, and so speech models stand to benefit from learning compressed input representations. 
Previous works have reduced sequence lengths to make training more tractable through fixed-length downsampling. 
However, phonemes are variable lengths.
Other work has shown promising results using phonemic representations and unsupervised term discovery from variable length sequences in MT and other domains, but as discrete units \cite{wilkinson2016subword,bansal2017towards,adams2016learning,kamper2016unsupervised,Dalmia2018SequenceBasedML,chung2018speech2vec}. 
Inspired by these works, we explore higher-level continuous speech embeddings for end-to-end speech translation.
Specifically, we use alignment methods to generate phoneme labels, and average consecutive frames with the same label to create phoneme-like feature vectors from variable numbers of frames.
We use the Fisher Spanish-English and low-resource Mboshi-French datasets.
We compare performance on the full Fisher dataset to smaller 
subsets as in \citet{bansal2018pre}.
As it is not possible to train a high-performing recognizer on many lower-resource tasks, we use a high-resource model applied cross-lingually to create phoneme labels for Mboshi.
We show significant performance improvements and reductions in training time under all conditions, demonstrating phoneme-informed speech representations are an effective and efficient tool for speech translation. 


\section{Method}

While frame-level Mel-frequency cepstral coefficient (MFCC) and filterbank features are informative, they create long, repetitive sequences which take recurrent models many examples to learn to model.
\begin{figure}[ht]
\centering
  \includegraphics[width=1\linewidth]{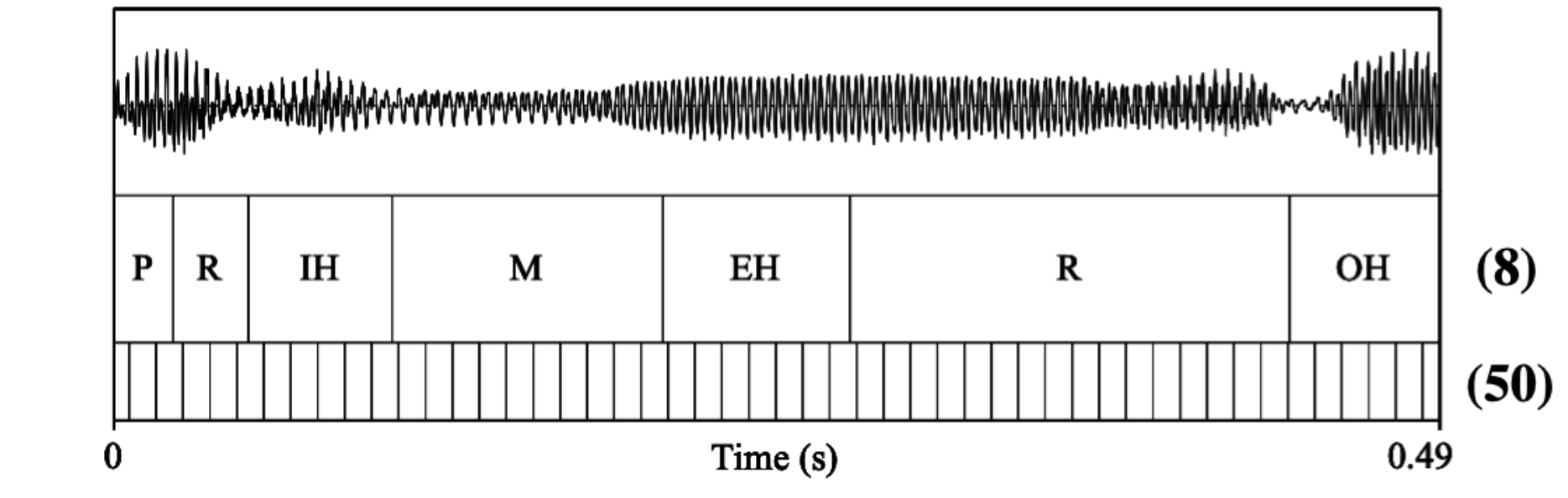}
  \caption{Example comparing number of frame-level features (50) to phoneme alignments (8). We saw an average reduction in sequence length of $\sim$80\%.}
  \label{comparison}
  \vspace{-1em}
\end{figure}
Higher-level representations like phonemes can create shorter, better-represented input sequences to improve training efficiency and model robustness.
Here, we average frame-level features within phoneme-like units to create one representation from a variable number of frames, using a trained speech recognizer and alignment. 

We extract 40-dimensional Mel filterbank features with per-speaker mean and variance normalization using Kaldi \cite{povey2011kaldi}.
Using an HMM/DNN system trained on the full Fisher Spanish dataset using the Kaldi \cite{povey2011kaldi} recipe for Fisher Spanish, we compute phoneme alignments using the triphone model (tri3a).
50 phoneme labels are used, including variants of silence, noise, and laughter. 
Within each utterance, we average the feature vectors for consecutive frames with the same label. 


The above method requires a recognizer with reasonable performance to perform alignment, not possible in low-resource conditions. 
Therefore, for Mboshi, we use a method that does not require language-specific data to generate a phoneme-like sequence.
Specifically, we apply a Connectionist Temporal Classification (CTC) model trained with 6000 hours of English data (notably, not a related language), as described in \citet{Dalmia2018SequenceBasedML} with the features from \citet{Dalmia2018DomainRF}.
To train this model, three frame-level features are spliced together, so output labels apply to a span of three frames.
Labels comprise a set of 40 phonemes and the CTC `blank' where the model is uncertain. 
The CTC `blank' transition label enables all frames to be aligned to a label. 
As above, we average the feature vectors for consecutive frames with the same label within an utterance.

\section{Model Architecture}
\label{sec:model}

As in \citet{bansal2018low}, we use a sequence-to-sequence architecture inspired by \citeauthor{weiss2017sequence} but modified to train within available resources; specifically, all models may be trained in less than 5 days on one GPU. 
We build an encoder-decoder model with attention in \texttt{xnmt} \citep{neubig2018xnmt} with 512 hidden units throughout. We use a 3-layer BiLSTM encoder.
We do not use the additional convolutional layers from \citeauthor{weiss2017sequence} and \citeauthor{bansal2018low} to reduce temporal resolution, but rather use network-in-network (NiN) projections from previous work in sequence-to-sequence ASR \cite{zhang2017very,sperber2018self} to get the same total $4\times$ downsampling in time. This gives the benefit of added depth with fewer parameters. 
We compare our performance to these two works in Section \ref{sec:baseline}.
We closely follow the LSTM/NiN encoder used in \citet{sperber2018self} for ASR and use the same training procedure, detailed in Appendix A.
We use an MLP attention with 1 hidden layer with 128 units and 64-dimensional target embeddings, though we use only 1 decoder hidden layer as opposed to 3 or 4 in previous works. 
All models use the same target preprocessing as previous work on this dataset: lowercasing and removing punctuation aside from apostrophes.


\section{Datasets}

\noindent\textbf{Spanish-English.} 
We use the Fisher Spanish speech corpus 
\cite{ldcfisherspanish}, which consists of 160 hours of telephone speech in multiple Spanish dialects split into 138K utterances, translated via crowdsourcing by \citet{post2013improved}.
We use the standard dev and test sets, each with $\sim$4k utterances. 
We do not use dev2.
Four reference translations are used to score dev and test.
\\

\noindent\textbf{Mboshi-French.} 
Mboshi is a Bantu language spoken in the Republic of Congo with $\sim$160k speakers. 
We use the Mboshi-French parallel corpus 
\cite{godard2017very} for our low-resource setting, which contains \textless5 hours of speech split into training and development sets of 4616 and 500 utterances respectively.
This corpus does not have a designated test set, so as in \citet{bansal2018pre} we removed 200 randomly sampled utterances from training for development data and use the designated development set as test.

\section{Results}

\subsection{Baseline}
\label{sec:baseline}

We first compare our model to previously reported end-to-end neural speech translation results on the Fisher Spanish-English task using frame-level features.
Table \ref{lit comparison} shows our results on the full training set with comparisons to \citet{weiss2017sequence} and \citet{bansal2018low}.
\begin{table}[ht]
\centering
\setlength\tabcolsep{5pt} 
\begin{tabular}{|c|cc|cc|cc|} 
\hline
\bf & \multicolumn{2}{c|}{\citeauthor{weiss2017sequence}} & \multicolumn{2}{c|}{\citeauthor{bansal2018low}} & \multicolumn{2}{c|}{Ours} \\ 
& \bf dev & \bf test & \bf dev & \bf test & \bf dev & \bf test \\ 
\hline
\bf BLEU & 46.5 & 47.3 & 29.5 & 29.4 & 32.4 & 33.7 \\
\hline
\end{tabular}
\caption{Single task end-to-end speech translation BLEU scores on full dataset.}
\label{lit comparison}
\vspace{-1em}
\end{table}
\citeauthor{weiss2017sequence}'s model is significantly deeper than ours, with 4 more encoder layers and 3 more decoder layers. After more than two weeks of expensive multi-GPU training, it reaches a 4-reference BLEU score of 47.3 on \texttt{test}.
We, like \citet{bansal2018low,bansal2018pre}, made modifications to our architecture and training schemes to train on a single GPU in approximately five days.
While \citeauthor{bansal2018low} use words on the target side to reduce time to convergence at a slight performance cost, we are able to use characters as in \citeauthor{weiss2017sequence} by having a still shallower architecture (2 fewer layers on both the encoder and decoder), which allows us to translate to characters with approximately the same training time per epoch they observe with words (${\sim}2$ hours).
We converge to a four-reference test BLEU of 33.7, showing 3-4 BLEU improvements over \citet{bansal2018low} on dev and test.
This demonstrates that our model has reasonable performance, providing a strong baseline before turning to our targeted task comparing input representations. 

\subsection{Frames vs Phonemes}


On our target task, we compare different subsets of the data to see how our method compares under different data conditions, using the full 160 hours as well as 40 and 20 hour subsets.
Table \ref{framevphone} shows our results using frame vs phoneme-level speech input.
When we use our phoneme-like embeddings, we see relative performance improvements of 13\% on all data sizes, or up to 5.2 BLEU on the full dataset. 
Further, in reducing source lengths by $\sim$80\%, training time is improved.
\begin{table}[b]
\centering
\setlength\tabcolsep{5pt} 
\begin{tabular}{|c|rr|rr|cc|} \hline
  & \multicolumn{2}{c|}{Frames} & \multicolumn{2}{c|}{Phonemes} & BLEU & Time \\
\bf Data & \bf dev & \bf test & \bf dev & \bf test & $\Delta$ & $\Delta$ \\ 
\hline
\bf Full & 32.4 & 33.7 & 37.6 & 38.8 & +5.2 & --67\% \\ 
\bf 40hr & 19.5 & 17.4 & 21.0 & 19.8 & +2.0 & --52\% \\ 
\bf 20hr &  9.8 &  8.9 & 11.1 & 10.0 & +1.2 & --65\% \\ 
\hline
\end{tabular}
\caption{Comparison of frame vs phoneme input on Spanish-English SLT, with average BLEU improvement and average reduction in training time.}
\label{framevphone}
\vspace{-1em}
\end{table}
We saw an average reduction in training time of 61\%, which for the full dataset means we were able to train our model in 39.5 hours rather than 118.2.

We compare our variable-length downsampling to fixed-stride downsampling by striding input frames. 
With a fixed stride of 2, performance decreases on 40 hours by $\sim$2 BLEU from 19.5 to 17.0 on dev and 17.4 to 15.6 on test.
With a fixed stride of 3, performance drops further to 13.7 and 11.8, respectively. 
By contrast, we saw improvements of +2 BLEU on 40 hours using our variable-length downsampling, though it lead to greater reductions in the number of input feature vectors. 
Clearly phoneme-informed reduction is far more effective than fixed schedule downsampling.

\subsection{Analysis}
To better understand our improvements, we target three points. 
Does making source and target sequence lengths more well-matched improve performance? To test we compare target preprocessing granularities. 
Second, reducing source lengths will impact both the encoder and attention. To investigate, we look at both encoder downsampling and ASR, where unlike MT, sequences are monotonic. 
Finally, we look at our low-resource case, Mboshi-French, where we must get phoneme labels from a cross-lingual source. 

\begin{table}[b]
\centering
\begin{tabular}{|c|c||c|c|c|c|c|} \hline
 Target & Target & \multicolumn{2}{c|}{Frames} & \multicolumn{2}{c|}{Phonemes} \\
 Preproc. & Length & dev & test & dev & test \\ \hline
\hline
chars   & 50.2 & 18.8 & \bf 17.3 & 20.0 & 18.4 \\ 
1k bpe  & 13.7 & \bf 19.5 & \bf 17.4 & \bf 21.0 & \bf 19.8 \\ 
10k bpe & 10.6 & 16.2 & 14.7 & 18.4 & 17.5 \\ 
words   & 10.4 & 16.4 & 14.6 & 18.2 & 17.4 \\ 
\hline
\end{tabular}
\caption{Comparing effects of target preprocessing with different sources on BLEU, Spanish-English 40hr}
\label{target preproc}
\vspace{-1em}
\end{table}
Previous work on sequence-to-sequence speech translation has used encoder downsampling of $4\times$, while $8\times$ is more common among sequence-to-sequence ASR systems \cite{zhang2017very}, motivated by reducing parameters and creating more one-to-one relationships between lengths of target sequence (typically characters) and the final encoder states to which the model attends. 
We use encoder downsampling of $4\times$, concatenating adjacent states after each layer. 
Table \ref{target preproc} shows target sequence lengths and results with different preprocessing. 
By averaging frames per local phoneme label in addition to encoder downsampling, source sequence lengths are further reduced on average by 79\%, yielding final encoder state lengths of 22, closest in length to 1k BPE targets (14) rather than characters (50). 
Given that the 1k BPE model performs best, it does appear that more similar source and target lengths boost performance.

\begin{figure}[ht]
\centering
  \includegraphics[width=1.0\linewidth]{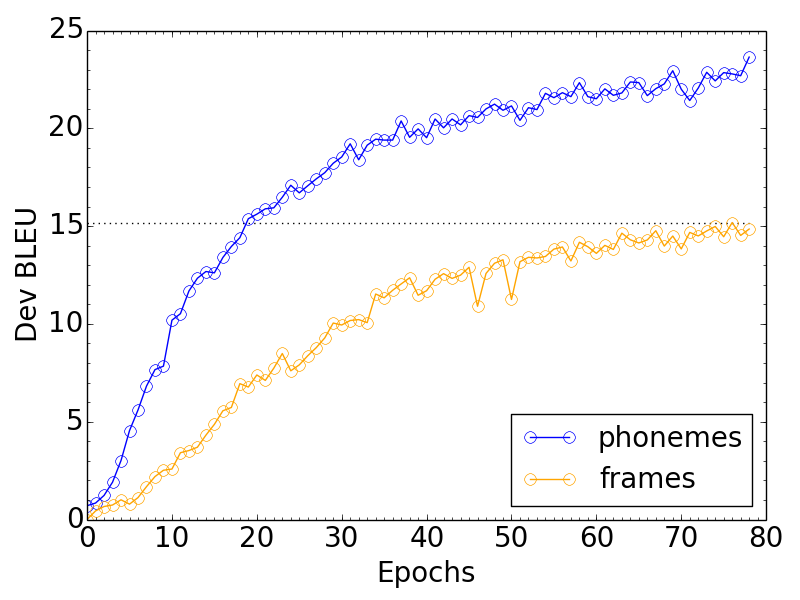}
  \caption{Dev BLEU over training with frames vs phonemes. 
  Single-reference BLEU on 1k lines of \texttt{dev}.}
  \label{training plot}
\end{figure}

For Spanish, we found that the mean number of frames per phone was 7.6, while the median was 6. 
Silence in this dataset skews these statistics higher; silence-marked frames account for 10.7\% of phone occurrences. 
Reducing multiple frames per phone to a single feature vector allows faster parameter optimization, as shown by improvements in early epochs in Figure \ref{training plot}.


We also compare the best phoneme models without encoder downsampling; with reduced sequence lengths, this becomes more tractable to train. 
On the full data, we see this improves our scores slightly, from 37.6 to 38.1 on dev and 38.8 to 39.2 on test. 
We see further improvements on 40 hours (22.4 dev \& 20.3 test), and on 20 hours, similar dev performance but slight improvements on test (10.3 dev \& 9.6 test). 
It is possible that with less data, the additional encoder parameters without downsampling do not receive enough updates to be well-trained. 


To test whether the approach is a generally more effective input representation or only an aid in the particularly complex learning task of speech translation where it helps to reduce the distance between inputs and outputs, we apply our method to ASR, where are alignments are monotonic.
We see similar levels of improvement, suggesting this approach produces generally more effective input representations: $\sim$18\% relative improvements on all three dataset sizes, or up to --9 absolute WER on 40 and 20 hours, as detailed in Table \ref{asr}. 
We note that \citet{weiss2017sequence} reported 25.7, 23.2 on dev and test, respectively, with a considerably larger network, which we are now able to match on test. 
\begin{table}[h]
\centering
\setlength\tabcolsep{4pt} 
\begin{tabular}{|c|rr|rr|cc|} \hline
  & \multicolumn{2}{c|}{Frames} & \multicolumn{2}{c|}{Phonemes} & WER & Time \\
\bf Data & \bf dev & \bf test & \bf dev & \bf test & $\Delta$ & $\Delta$ \\ 
\hline
\bf Full & 33.4 & 30.0 &  28.0 & 23.4 &  --6.0 & --43\% \\ 
\bf 40hr & 44.8 & 46.7 &  36.6 & 36.6 &  --9.2 & --40\% \\ 
\bf 20hr & 56.3 & 59.1 &  48.2 & 49.1 &  --9.1 & --50\% \\ 
\hline 
\end{tabular}
\caption{Comparison of frame vs phoneme input on Spanish ASR, with average reduction in WER and average reduction in training time.}
\label{asr}
\end{table}
We note that this neural model also outperforms the Kaldi models; the Kaldi model using the tri3a alignments we use for phoneme boundaries yields 45.7 dev WER, and using more sophisticated alignment models, achieves 29.8.

On our low-resource Mboshi task, we do not have enough data to train a high-quality recognizer to produce phoneme alignments. Instead, we use a model from an unrelated language (English) applied cross-lingually.
With small training and evaluation sets, scores are less stable and changes must be taken with a grain of salt. 
We see very low scores with frames, but still see improvements with phonemes, though the labels were produced by an English model. 
\citet{bansal2018pre} reported 3.5 BLEU using frames, which they improved to 7.1 by pretraining their encoder with 300 hours of English and decoder with 20 hours of French. 
Creating phoneme-level embeddings, we are able to get similar levels of improvement without training the network on more data, though we use an unadapted foreign language model.

\begin{table}[h]
\centering
\setlength\tabcolsep{6pt} 
\begin{tabular}{|l|rr|rr|} \hline
  & \multicolumn{2}{c|}{Frames} & \multicolumn{2}{c|}{Phonemes} \\
\bf Data & \bf dev & \bf test & \bf dev & \bf test \\ 
\hline
Mboshi (chars)   & 0.0 & 0.0 &   5.2 & 3.6 \\ 
Mboshi (1k bpe)  & 2.3 & 1.4 &   7.0 & 5.6 \\ 
Mboshi (words)  & 1.8 & 1.4 &    7.8 & 5.9 \\ 
\hline
\end{tabular}
\caption{Comparison of frame vs phoneme input on Mboshi-French SLT. Mboshi phoneme labels produced with English CTC phoneme recognizer.} \label{mboshi}
\end{table}



While LSTM-based sequence-to-sequence models are able to learn from long, redundant sequences, we show that they learn more efficiently and effectively across multiple data conditions when given sequences reduced using phoneme boundaries.
This is evidenced by our improvements across all data sizes, and significant improvements in early epochs, shown in Figure \ref{training plot}. 

We compared two methods for alignment, an HMM-based model and a CTC-based model, the first applied monolingually and the second cross-lingually. 
The CTC model yields blank alignments for some frames, reducing the range of frames to be averaged, though the center of mass often remains the same. 
We hypothesize that this does not greatly impact results, and previous work has explored using the middle HMM state for alignments rather than all \cite{stuker2003multilingual}, but this would benefit from a more robust comparison. 
As well, a deeper comparison of monolingual versus cross-lingual alignments applied to a greater number of test languages would be beneficial. 

\section{Conclusion}

Previous work on end-to-end speech translation has used frame-level speech features. 
We have shown that a na{\"i}ve method to create higher-level speech representations for translation can be more effective and efficient than traditional frame-level features. 
We compared two input representations for two unrelated languages pairs, and a variety of differently-resourced conditions, using both a supervised alignment method and a cross-lingual method for our low-resource case. 
Our method does not introduce additional parameters: we hope to motivate future work on learning speech representations, with continued performance on lower-resource settings if additional parameters are introduced.


\paragraph{Acknowledgements}
We would like to thank the anonymous reviewers for their helpful comments.

\bibliography{bibliography}
\bibliographystyle{acl_natbib}

\appendix

\section{Appendix. LSTM/NiN Encoder and Training Procedure Details}
\label{sec:appendix}

\subsection{Encoder Downsampling Procedure}
\citet{weiss2017sequence} and \citet{bansal2018low} use two strided convolutional layers atop three bidirectional long short-term memory (LSTM) \cite{Hochreiter1997} layers to downsample input sequences in \textbf{time} by a total factor of 4.
\citet{weiss2017sequence} additionally downsample \textbf{feature} dimensionality by a factor of 3 using a ConvLSTM layer between their convolutional and LSTM layers. 
This is in contrast to the pyramidal encoder \cite{chan2016listen} from sequence-to-sequence speech recognition, where pairs of consecutive layer outputs are concatenated before being fed to the next layer to halve the number of states between layers. 

To downsample in time we instead use the LSTM/NiN model used in \citet{sperber2018self} and \citet{zhang2017very}, which stacks blocks consisting of an LSTM, a network-in-network (NiN) projection, layer batch normalization and then a ReLU non-linearity. 
NiN denotes a simple linear projection applied at every timestep, performing downsampling by a factor of 2 by concatenating pairs of adjacent projection inputs.
The LSTM/NiN blocks are extended by a final LSTM layer for a total of three BiLSTM layers with the same total downsampling of 4 as \citet{weiss2017sequence} and \citet{bansal2018low}. 
These blocks give us the benefit of added depth with fewer parameters.

\subsection{Training Procedure}
We follow the training procedure from \citet{sperber2018self}.
The model uses variational recurrent dropout with probability 0.2 and target character dropout with probability 0.1 \cite{gal2016theoretically}.
We apply label smoothing \cite{szegedy2016rethinking} and fix the target embedding norm to 1 \cite{nguyen2017improving}.
For inference, we use a beam size of 15 and length normalization with exponent 1.5.
We set the batch size dynamically depending on the input sequence length such that the average batch size was 36.
We use Adam \cite{kingma2014adam} with initial learning rate of 0.0003, decayed by 0.5 when validation BLEU did not improve over 10 epochs initially and 5 epochs after the first decay.
We do not use L2 weight decay or Gaussian noise, and use a single model replica.
All models use the same preprocessing as previous work on this dataset: lowercasing and removing punctuation aside from apostrophes. 
We use input feeding \cite{Luong2015b}, and we exclude utterances longer than 1500 frames to manage memory requirements.

\end{document}